\documentclass[runningheads]{llncs}

\usepackage[preprint,year=2024,ID=6029]{eccv}
\usepackage{eccvabbrv}

\usepackage{graphicx}
\usepackage{booktabs}

\usepackage[accsupp]{axessibility}  % Improves PDF readability for those with disabilities.

\usepackage{multirow}
\usepackage{bbding}

\usepackage[pagebackref,breaklinks,colorlinks,citecolor=eccvblue]{hyperref}
\usepackage{orcidlink}

\begin{document}

\title{SGDM: Static-Guided Dynamic Module Make Stronger Visual Models} 

\titlerunning{}

\author{Wenjie Xing\inst{1,2}\orcidlink{0000-0003-1163-5661} \and
Zhenchao Cui\inst{1,2} \and Jing Qi\inst{1,2}}

\authorrunning{W.~Xing et al.}

\institute{School of Cyber Security and Computer Hebei University, Baoding, 071002, China \and
Hebei Machine Vision Engineering Research Center Hebei University, Baoding, 071002, China}

\maketitle

\begin{abstract}
  The spatial attention mechanism has been widely used to improve object detection performance. However, its operation is currently limited to static convolutions lacking content-adaptive features. This paper innovatively approaches from the perspective of dynamic convolution. We propose \emph{Razor Dynamic Convolution (RDConv)} to address the two flaws in dynamic weight convolution, making it hard to implement in spatial mechanism: 1) it is computation-heavy; 2) when generating weights, spatial information is disregarded. Firstly, by using \emph{Razor Operation} to generate certain features, we vastly reduce the parameters of the entire dynamic convolution operation. Secondly, we added a spatial branch inside RDConv to generate convolutional kernel parameters with richer spatial information. Embedding dynamic convolution will also bring the problem of sensitivity to high-frequency noise. We propose the \emph{Static-Guided Dynamic Module (SGDM)} to address this limitation. By using SGDM, we utilize a set of asymmetric static convolution kernel parameters to guide the construction of dynamic convolution. We introduce the mechanism of shared weights in static convolution to solve the problem of dynamic convolution being sensitive to high-frequency noise. Extensive experiments illustrate that multiple different object detection backbones equipped with SGDM achieve a highly competitive boost in performance(\eg, +4\% mAP with YOLOv5n on VOC and +1.7\% mAP with YOLOv8n on COCO) with negligible parameter increase(\ie, +0.33M on YOLOv5n and +0.19M on YOLOv8n).
\end{abstract}

\section{Introduction}
\label{introduction}
Object detection is a pivotal component within computer vision, entailing the recognition and precise localization of distinct entities within images or video streams. Object detection's rapid development and application mark its substantial influence and widespread applicability across various artificial intelligence sectors. In recent years, the advent of attention mechanisms, exemplified by the Convolutional Block Attention Module (CBAM) \cite{woo2018cbam}, has significantly augmented the capabilities of object detection frameworks, thereby enhancing model performance and establishing new benchmarks in the field. The latest, such as RFAConv\cite{zhang2023rfaconv}, not only focuses on the spatial information in the feature map but also pays attention to weight sharing in large convolutional kernels, further enhancing the effectiveness of attention mechanisms.

At the same time, Dynamic weight convolutions \cite{sun2021gaussian, ding2021dynamic, yang2019condconv, chen2020dynamic, zhang2020dynet, zhou2021decoupled, ma2020weightnet, li2021revisiting} has gradually emerged due to excellent performance. Dynamic convolution exhibits spatial-anisotropy and content-adaptive properties\cite{hu2022bi}, resulting in stronger feature extraction capability. 

The above characteristics of dynamic convolution make it easy for people to associate it with spatial. However, some unresolved issues exist in embedding dynamic convolution in spatial attention mechanisms.

Firstly, due to dynamic weight convolutions requiring combining multiple convolutional kernel parameters, their parameter quantity is proportional to the number of kernels. Therefore, networks that apply with dynamic weight convolutions have times more parameters, which undoubtedly brings about issues such as model complexity and slow inference, which violates the principle of attention mechanism plug and play.

Secondly, the squeeze-and-excitation module (SE) \cite{hu2018squeeze} has become the de facto standard for generating dynamic weights \cite{yang2019condconv, chen2020dynamic, li2022omnidimensional}. Besides, dynamic weight convolution is mainly aimed at tasks such as classification. However, spatial information in images is crucial for dense tasks like object detection. In SE, spatial information is compressed by global average pooling. It will inevitably destroy the spatial information in feature maps, thus leading to sub-optimal results in object detection tasks.

Thirdly, as dynamic convolution calculates the weight for each image separately, for high-frequency information in some images, such as noises and edges, dynamic convolution sometimes treats noise as an object or causes errors in edge recognition.

Therefore, in this paper, we propose \emph{RDConv} and \emph{SGDM}. Firstly, we take a different perspective to reduce parameters. We introduce \emph{Razor Operation} to reduce redundant features that enter the RDConv, thus solving the problem from the input end. Secondly, we add a spatial branch internally to optimize the parameter generation inside RDConv. In this way, RDConv can obtain more abundant spatial information. Thirdly, through the homogeneous asymmetric structure of SGDM, we fuse the weight of dynamic convolution with that of asymmetric static convolution. Due to the shared weight characteristic of static convolution, it is less sensitive to high-frequency noise. Therefore, we use the parameters of static convolutional kernels to guide the parameter generation in RDConv, enhancing the overall anti-interference capability of SGDM.

We experiment with the proposed SGDM regarding qualitative and quantitative evaluations with mainstream visual models on multiple datasets. Extensive results demonstrate that SGDM achieves highly competitive improvements (\ie, object detection on COCO \cite{lin2014microsoft}: +2.6\% AP with YOLOv5n, +1.7\% AP with YOLOv8n) while bringing a negligible parameter increase and computational cost.

\begin{figure*}[t]
  \centering
  \begin{subfigure}{0.35\linewidth}
    \includegraphics[width=\textwidth]{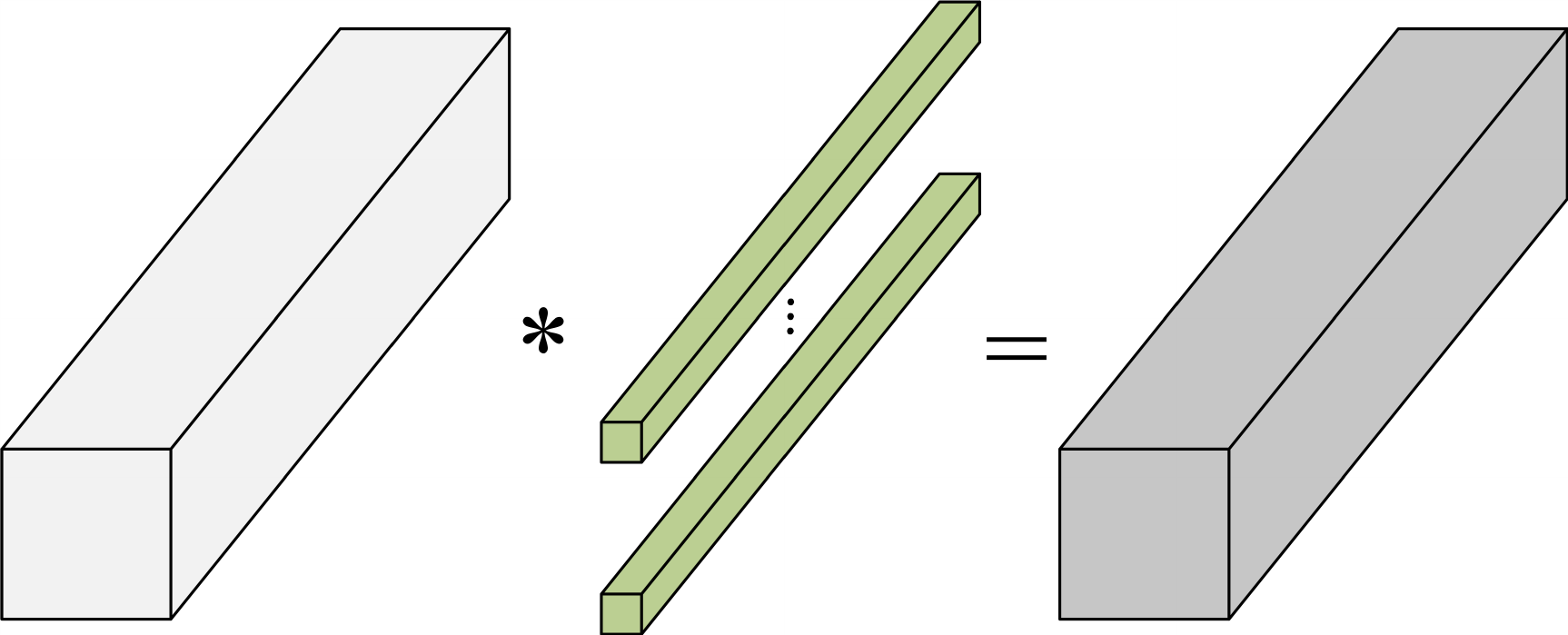} 
    \caption{Typical Dynamic Convolution.}
    \label{fig1a}
  \end{subfigure}
  \hfill
  \begin{subfigure}{0.6\linewidth}
    \includegraphics[width=\textwidth]{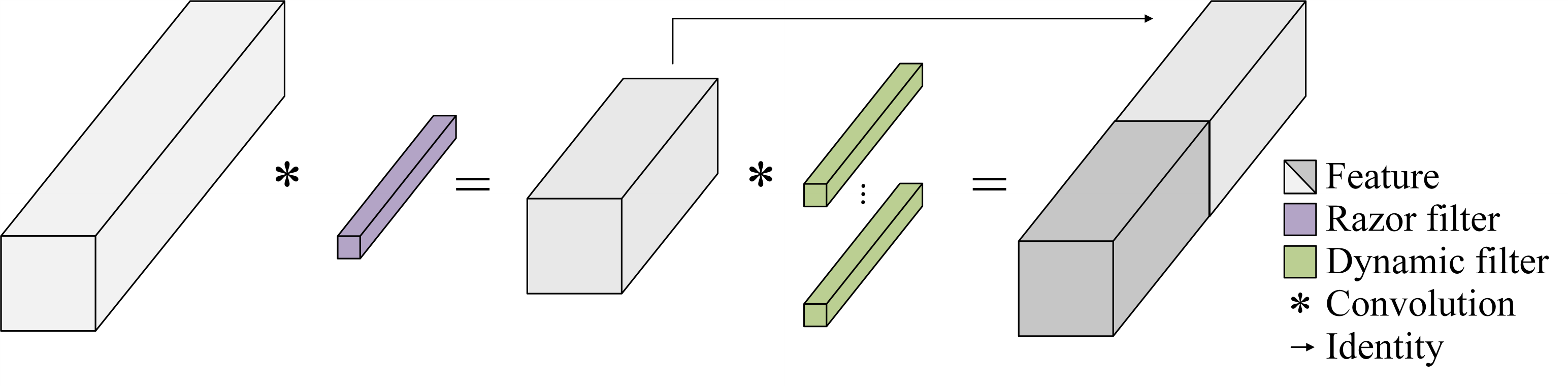}  
    \caption{Razor Dynamic Convolution.}
    \label{fig1b}
  \end{subfigure}
  \caption{Illustration of typical dynamic convolution and our RDConv. The former calculates
dynamic weights by all features, while the latter only operates on intrinsic features.}
  \label{fig1}
\end{figure*}

\section{Related Work}
\subsection{Object Detection Models}
The object detection task has always been a hot direction in machine learning. Among them, some models are particularly eye-catching. YOLOv5 \cite{Jocher_YOLOv5_by_Ultralytics_2020} provides high accuracy, low computational complexity, and high speed, making it widely applicable in industrial scenarios. Moreover, YOLOv6 \cite{li2023yolov6} includes bidirectional concatenation as the feature map fusion module in the network's neck. Then, YOLOv8 \cite{Jocher_YOLO_by_Ultralytics_2023} introduces the C2f module as the backbone stage module and incorporates an anchor-free detection head. Meanwhile, it uses a task-aligned assigner that combines classification and localization information. On the other hand, EfficientViT combines the efficiency of EfficientNet with the global context-capturing capabilities of Vision Transformers, aiming for parameter efficiency and enhanced performance. Pure CNN networks like ConvNeXtv2\cite{woo2023convnext} and FasterNet\cite{chen2023run} have also achieved impressive results.

\subsection{Attention Mechanisms}
As a technique to improve performance, the attention mechanism enables models to focus on essential features. The SE block \cite{hu2018squeeze} is a notable example of channel attention. It calculates weights for each channel by compressing features to gather global channel information. Additionally, the Convolutional Block Attention Module (CBAM) \cite{woo2018cbam} combines channel attention and spatial attention. It can be easily integrated into convolutional neural networks as a plug-and-play module. Despite the success of SE and CBAM, they tend to lose significant information during feature compression. The lightweight coordinate attention (CA) \cite{hou2021coordinate} was introduced to address this issue. Furthermore, RFAConv \cite{zhang2023rfaconv} reevaluated the spatial attention mechanism by considering convolutional kernel parameter sharing.

\subsection{Dynamic weight convolutions}
For CNN, implementing dynamic weight convolutions to adapt a neural network's weights to samples has shown potential for improving model capacity and generalization. Dynamic weight convolutions \cite{yang2019condconv, chen2020dynamic} generates multiple sets of parameters for each convolution kernel and then weights them using an attention mechanism based on the input. However, using dynamic weight convolution can pose challenges. Firstly, it can be sensitive to high-frequency noise in images. Bi-volution \cite{hu2022bi} introduces a dual-branch structure to leverage the complementary properties of static/dynamic convolution effectively. Secondly, dynamic convolution may have more parameters and computations than static convolution. Several techniques have been proposed to address this, including DDF \cite{zhou2021decoupled}, DCD \cite{li2021revisiting}, and ODConv \cite{li2022omnidimensional}.

\section{Method}
In this section, we first review standard and dynamic convolution via a general formulation. Then, we formalize the common spatial attention mechanism. After that, we describe the formulation of our RDConv and clarify its properties. Finally, we introduce the specific composition of SGDM.

\subsection{Dynamic Convolution Preliminaries}
Given an input feature representation $x \in \mathbb{R}^{c\times n}$ with $c$ channels and $n$ pixels ($n=h\times w$, $h$ and $w$ are the width and height of the feature map); the standard convolution operation at $i^{th}$ pixel can be written as a linear combination of input features around $i^{th}$ pixel:
\begin{equation}
\label{eq1}
    X^{out}_{(.,i)}=\sum_{j \in \Omega(i)}W[\mathrm{p}_i - \mathrm{p}_j]X_{(.,j)}+\mathrm{b},
\end{equation}
where $X_{(.,j)} \in \mathbb{R}^{c_{in}}$ denotes the feature vector at $j^{th}$ pixel; $X^{out}_{(.,i)} \in \mathbb{R}^{c_{out}\times n}$ denotes $i^{th}$ pixel output feature vector. $\Omega(i)$ denotes the $k \times k$ convolution window around $i^{th}$ pixel. $W \in \mathbb{R}^{c_{out} \times c_{in} \times k \times k}$ is a $k \times k$ convolution filter, $W[\mathrm{p}_i - \mathrm{p}_j] \in \mathbb{R}^{c_{out} \times c_{in}}$ is the filter at position offset between $i$ and $j^{th}$ pixels: $[\mathrm{p}_i - \mathrm{p}_j] \in {\left(-\frac{k-1}{2}, -\frac{k-1}{2}\right), \left(-\frac{k-1}{2}, -\frac{k-3}{2}\right), \ldots, \left(\frac{k-1}{2}, \frac{k-1}{2}\right)}$ where $\mathrm{p}_i$ denotes 2D pixel coordinates. $\mathrm{b} \in \mathbb{R}^{c_{out}}$ denotes the bias vector. In standard convolution, the same filter $W$ is shared across all samples, and filter weights are agnostic to input features.

Unlike standard convolution, dynamic convolution leverages separate network branches to generate unique filters for each sample. The spatially-invariant filter $W$ in \cref{eq1} becomes the spatially-varying filter $D_i \in \mathbb{R}^{c_{out} \times c_{in} \times k \times k}$ in this case. In other words, it uses a linear combination of $n$ convolutional kernels weighted dynamically with an attention mechanism, making convolution operations input-dependent. From the view of input and output features, dynamic convolution operations can be defined as:
\begin{equation}
\label{eq2}    
    X^{out}=(\alpha_{d1}D_1+ \ldots + \alpha_{dn}D_n)*X,
\end{equation}            
where $X \in \mathbb{R}^{c_{in} \times h \times w}$ denotes the input features and $X^{out} \in \mathbb{R}^{c_{out} \times h \times w}$ denotes the output features, respectively; $D_i$ denotes the $i^{th}$ convolutional kernel's weight; $\alpha_{di} \in \mathbb{R}$ is the attention scalar for weighting $D_i$, which is computed by an attention function $\pi_{di}(\cdot)$ conditioned on the input features; $*$ denotes the convolution operation.

\subsection{Spatial Attention Mechanism Preliminaries}
Currently, the spatial attention mechanism uses the attention map obtained through learning to highlight the importance of each feature. The spatial attention mechanism can be expressed as follows:
\begin{equation}
\begin{aligned}
\label{eq3}    
    W_{[1, 2, \cdots, n]} = A_{[1, 2, \cdots, n]} * X_{[1, 2, \cdots, n]}
\end{aligned}
\end{equation}
where $W_i$ represents the value obtained after the weighting operation. $A_i$ and $X_i$ represent the learned attention map at different positions and the input feature map, respectively. $n$ is the product of the length and width of the feature map, which represents the total number of pixel values.

In the spatial attention mechanism, $A_i$ is usually generated by convolution and pooling operations. Taking the Spatial Attention Module in CBAM as an example, its formulation is shown as follows:
\begin{equation}
\begin{aligned}
\label{eq4}    
   X^{out} = \sigma(\operatorname{F}([\operatorname{AP(X)}; \operatorname{MP}(X)]))
\end{aligned}
\end{equation}
where $\sigma$ denotes the sigmoid, $\operatorname{F}$ represents convolution operation with a $7\times7$ kernel size, and $\operatorname{AP}$ and $\operatorname{MP}$ respectively represent Average Pooling and Max pooling.

From \cref{eq4}, we can see that the essence of spatial attention is to assign a weight to the output, telling the network where to pay more attention. However, this operation is still content-independent. After the network training is completed, spatial attention will only focus on specific areas for all inputs. 

\begin{figure*}[t]
  \centering
  \begin{subfigure}{0.5\linewidth}
    \includegraphics[width=\textwidth]{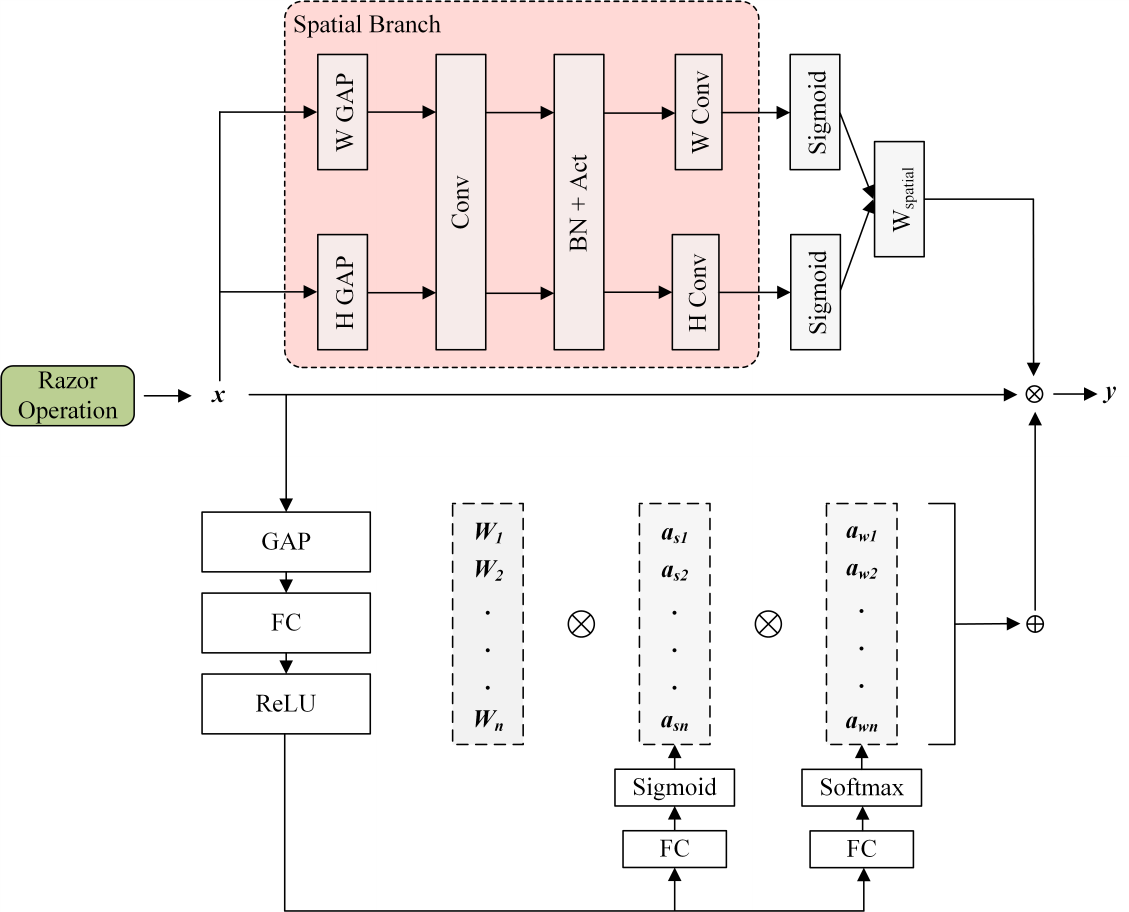} 
    \caption{The overall flowchart of RDConv after Razor Operation.}
    \label{fig2a}
  \end{subfigure}
  \hfill
  \begin{subfigure}{0.45\linewidth}
    \includegraphics[width=\textwidth]{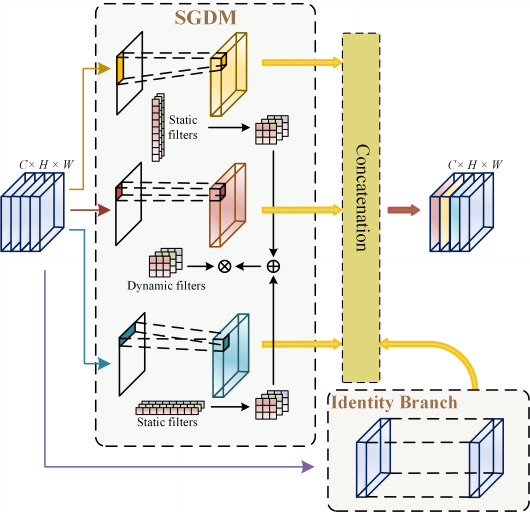}  
    \caption{Overview of the Static-Guided Dynamic Module.}
    \label{fig2b}
  \end{subfigure}
  \caption{Illustration of the composition of RDConv and the design of SGDM module. RDConv effectively solves some problems of dynamic convolution through \emph{Razor Operation} and Spatial Branch. The SGDM module is a plug-and-play module while applying RDConv, achieving seamless performance improvement for the visual models.}
\end{figure*}

\subsection{Razor Dynamic Convolution}
Firstly, we analyze the attention function $\pi_{di}(\cdot)$ of dynamic convolution: its main role is to find the differences between features across channels and attach a weight to them. Mathematically,  $\pi_{di}(\cdot)$ can be defined as:
\begin{equation}
\label{eq5}    
    \pi_{di}(\cdot) = \sigma(X_i, \phi(X_i; \theta_i)),
\end{equation}
where $W_i$ represents each dynamic convolution's parameters ($i$ usually sets to 4), and $X_i \in \mathbb{R}^{c_{in} \times h \times w}$ represents features entering $\pi_{di}(\cdot)$ to calculate $ W_i$. $\sigma$ denotes the Sigmoid activate function, $\phi$ denotes the sub-network used to predict the kernel parameters from $X_i$, and $\theta_i$ are the weights of $\phi$. From \cref{eq3}, we can see that for each calculation of dynamic convolution parameters, the attention function uses all the inputs, typically $ c_{in} \times h \times w$ values. During $\pi_{di}(\cdot)$, the required number of FLOPs can be calculated as $ c_{in} \times c_{in} \times h \times w$, which is square related to input channels. In common CNN networks, $ c_{in} $ is generally very large (\eg, 512 or 1024).

So, the parameters of dynamic convolution can be directly determined by the number of channels entering $\pi_{di}(\cdot)$. Moreover, we know redundant features can be excluded from calculating dynamic attention weights. It is inefficient and unnecessary to input all feature maps into $\pi_{di}(\cdot)$. So compared to typical dynamic convolution shown in \cref{fig1a}, we first compress the channels by \emph{Razor Operation} to reduce the number of features entering $\pi_{di}(\cdot)$ as shown in \cref{fig1b}. Then, the cheap operation is used to generate the remaining features directly. Compared with the typical dynamic convolution, RDConv dramatically reduces the amount of calculation by performing dynamic operations on only a few intrinsic features. Our \emph{Razor Operation} can be expressed as the following equation:
\begin{equation}
\label{eq6}    
    \pi_{di}(\cdot) = \sigma(\delta(X_i, (X_i; \theta_i))),
\end{equation}
where $\delta$ is a channel compress operation that prioritizes intrinsic features and $\sigma$ denotes the same Sigmoid activate function. Assuming $r$ is the proportion of feature reduction, after adding \emph{Razor Operation}, RDConv can be expressed as:
\begin{equation}
\label{eq7}    
    X^{out}=r(\alpha_{d1}D_1+ \ldots + \alpha_{dn}D_n)X + (1-r)X
\end{equation}

After dealing with the problem of high parameters, we are about to tackle the issues of insufficient spatial information acquisition. 

We add a spatial branch to obtain more spatial information for weight generation, as shown in \cref{fig2a}. The output of the spatial branch is also a weight, which we multiply with the dynamic convolution output to generate convolution weights better. 

Since RDConv focuses on reducing parameters, introducing spatial information with a separate spatial branch could potentially increase parameters. So, as shown in \cref{fig2a}, we decouple height and width dimensions to calculate the spatial attention separately. In this way, we reduce the parameter amount by about $1/3$ (\eg, an original $3\times3$ convolutional kernel is replaced by $1\times3$ and $3\times1$ convolutional kernels, and the parameter amount decreases from $3\times 3 = 9$ to $1\times 3 + 3\times 1 = 6$). 

Specifically, H-GAP is a Global Average Pooling in the height dimension. W-GAP is a Global Average Pooling in the width dimension. H-Conv and W-Conv are vital components for obtaining spatial information, and the design of their convolutional kernel size is crucial as it represents a balance between the richness of spatial information obtained by the entire spatial branch and computational complexity. In this regard, we conducted ablation experiments in \cref{Ablation studies} to confirm parameter selection and ultimately chose a kernel size of $1 \times 15$ and $15\times 1$. This way, RDConv acquires abundant spatial information through separate feature extraction without increasing too many parameters. 

\subsection{Static-Guided Dynamic Module}
After designing RDConv, we have solved two problems in dynamic convolution: high parameter volume and lack of spatial information. Another issue is the sensitivity to high-frequency noise brought by the content-adaptive feature. To address this, we design the SGDM using shared weights in static convolution kernels to guide the generation of convolution kernel parameters in RDConv during filtering processes to not overly focus on sudden anomalies and thus resolve the sensitivity to high-frequency noise.

In \cref{fig2b}, we present the detailed design of SGDM. Specifically, for input $X \in \mathbb{R}^{c_{in} \times h \times w}$, to avoid excessive computational load caused by SGDM, making it unable to become a plug n' play module, we consider separating features along the channel dimension. Specifically, we first divide the feature into four groups, namely $X_{rd}, X_h, X_w,$ and $X_{id}$ with segmentation ratios of $[r, r, r, 1-3r]$. $X_{rd}$ represents the feature entering RDConv; $X_h$ and $X_w$ represent the input features for a set of asymmetric static convolutions with kernel sizes of $k_s \times 1$ and $1 \times k_s$, respectively; $X_{id}$ represents the identity mapping from the original feature without further processing. The four parallel outputs of SGDM can be expressed as follows:
\begin{equation}
\left\{
\begin{aligned}
\label{eq8}    
    X^{\prime}_{rd} &= \operatorname{RDConv}^{rc_{in} \rightarrow rc_{in}}_{k_d \times k_d}(X_{rd}), \\
    X^{\prime}_{h} &= \operatorname{StaticConv}^{rc_{in} \rightarrow rc_{in}}_{k_s \times 1}(X_{h}), \\
    X^{\prime}_{w} &= \operatorname{StaticConv}^{rc_{in} \rightarrow rc_{in}}_{1 \times k_s}(X_{w}), \\
    X^{\prime}_{id} &= \operatorname{Identity}^{(1-3r)c_{in} \rightarrow (1-3r)c_{in}}(X_{id})
\end{aligned}
\right.
\end{equation}
where $k_d$ denotes the square kernel size we use in RDConv, which is set to 3 by default; $k_s$ represents the strip kernel size used in asymmetric convolution and is set as nine by default, so that the parameters of two static convolutional kernels can be easily reshaped from $9 \times 1$ and $1 \times 9$ to $3\times3$, thus guiding the generation of internal kernel parameters in RDConv. $rc_{in}\rightarrow rc_{in}$ represents channel change after entering each branch. In SGDM, we set $r=0.25$ by default. At last, the final output is concatenated along the channel dimension and can be described as follows:
\begin{equation}
\label{eq9}    
    X^{\prime} = \operatorname{Cat}(X^{\prime}_{rd}, X^{\prime}_{h}, X^{\prime}_{h}, X^{\prime}_{id})
\end{equation}

At the same time, we denote the weights generated by the static convolutional kernel as $W_h$ and $W_w$. They were initially both in vertical bar shape. We use a simple mathematical transformation to generate a static convolutional kernel parameter with the same shape as dynamic convolutional kernel parameters.

Due to the shared weight characteristic of static convolution, they have stronger robustness against abnormal noise in the image than dynamic convolution. Therefore, SGDM can achieve content-adaptiveness while being able to resist high-frequency noise. Specifically, the parameter fusion process is as follows:

\begin{equation}
\begin{aligned}
\label{eq10}    
    W_{rd}^{\prime} = W_{rd} * (\psi(W_h) + \psi(W_w))
\end{aligned}
\end{equation}
where $\psi$ is the reshape operation mentioned in \cref{eq10}. First, we sum the generated static convolutional weights and then take the dot product with the dynamic convolutional kernel parameters to obtain the final parameters.

\begin{table}[h]
\caption{Comparison of training recipes among YOLO series, n represents the nano model, s represents the relatively large model.}
\label{table1}
\centering
\begin{tabular}{lccc}
\toprule
Params. & YOLOv5 & YOLOv6 & YOLOv8\\
\midrule
Epochs & 300 & 300 & 300\\
Batch size & 72(n)/48(s) & 128(n)/72(s) & 48(n)/36(s)\\
Image size & $640^2$ & $640^2$ & $640^2$\\
Optimizer & SGD & SGD & SGD\\
LR & 1e-2 & 1e-2 & 1e-2\\
LR decay & 5e-4 & 5e-4 & 5e-4\\
Momentum & 0.937 & 0.937 & 0.937\\
Label smoothing & \XSolidBrush & 0.0 & 0.0 \\
Close mosaic & \XSolidBrush & 10 & 10 \\
\bottomrule
\end{tabular}
\end{table}

\section{Experiments}
We perform comparative experiments on the object detection track of the VOC and MS-COCO datasets. We use 16,551 images with 20 categories in the VOC 2007 trainval and VOC 2012 trainval datasets for training and the VOC 2012 test dataset for validation. MS-COCO 2017 contains 118,287 training images and 5,000 validation images with 80 object classes. We use the code frameworks of YOLOv5, YOLOv6, and YOLOv8 for YOLO series models. To verify the effectiveness of SGDM across different architectures, we also embed modules into works such as \cite{liu2023efficientvit, chen2023run, woo2023convnext}. For experiments on the VOC dataset, YOLOv5, YOLOv6, and YOLOv8 all use open-source models from the Ultralytics \cite{Jocher_YOLO_by_Ultralytics_2023} library to ensure controllable parameter quantities during the experiment process. For experiments on the MS-COCO dataset, we used the respective open-source libraries of YOLOv5 \cite{Jocher_YOLOv5_by_Ultralytics_2020} to compare results with RFAConv. We compared CA and CBAM \cite{woo2018cbam} and RFAConv \cite{zhang2023rfaconv} on multiple backbones. At the same time, to verify the effectiveness of RDConv, we compared CondConv \cite{yang2019condconv}, DyConv \cite{chen2020dynamic}, DCD \cite{li2021revisiting}, and ODConv \cite{li2022omnidimensional}.

\begin{table*}[h]
\caption{Results on the VOC 2012 test set compared to different attention mechanisms on multiple backbones. The best results are bolded.}
\label{table2}
\centering
\resizebox{\linewidth}{!}{
\begin{tabular}{lcccccccc}
\toprule
\multirow{2}*{Backbones} & Params(M) & FLOPs(G) & $\text{AP}_{50}(\%)$ & $\text{AP}(\%)$ & Params(M) & FLOPs(G) & $\text{AP}_{50}(\%)$ & $\text{AP}(\%)$ \\
\cmidrule(lr){2-5}\cmidrule(lr){6-9} & \multicolumn{4}{c}{YOLOv5n} & \multicolumn{4}{c}{YOLOv5s} \\
\midrule
Baseline & 2.5 & 7.2 & 77.9 & 56.3 & 9.2 & 24.2 & 82.6 & 61.7 \\
+ CBAM & 2.6 & 7.3 & 77.3 & 56.1 & 9.5 & 24.5 & 82.3 & 61.7 \\
+ CA & 2.7 & 7.9 & 78.0 & 56.8 & 9.2 & 24.3 & 82.7 & 62.5 \\
+ RFAConv & 3.3 & 8.8 & 78.9 & 58.1 & 12.3 & 30.3 & 83.4 & 63.5 \\
+ SGDM & 2.8 & 8.1 & $\mathbf{79.9}$ & $\mathbf{60.3}$ & 9.8 & 25.1 & $\mathbf{83.8}$ & $\mathbf{64.9}$ \\
\midrule \midrule
\quad & \multicolumn{4}{c}{YOLOv6n} & \multicolumn{4}{c}{YOLOv6s} \\
\midrule
Baseline & 4.5 & 13.1 & 79.3 & 59.5 & 16.5 & 44.9 & 83.6 & 64.6 \\
+ CBAM & 4.5 & 13.1 & 80.4 & 60.1 & 16.5 & 44.9 & 83.5 & 64.8 \\
+ CA & 4.5 & 13.1 & 79.1 & 59.2 & 16.5 & 44.9 & 83.3 & 64.5 \\
+ RFAConv & 4.7 & 13.6 & 79.6 & 60.3 & 17.3 & 46.5 & 83.8 & 65.2 \\
+ SGDM & 4.5 & 13.2 & $\mathbf{80.8}$ & $\mathbf{61.1}$ & 16.6 & 45.1 & $\mathbf{83.9}$ & $\mathbf{65.4}$ \\
\midrule
\midrule
\quad & \multicolumn{4}{c}{YOLOv8n} & \multicolumn{4}{c}{YOLOv8s} \\
\midrule
Baseline & 3.1 & 8.7 & 79.4 & 59.3 & 11.2 & 28.8 & 83.1 & 63.5 \\
+ CBAM & 3.2 & 8.9 & 78.9 & 58.9 & 11.5 & 29.1 & 82.3 & 63.0 \\
+ CA & 3.2 & 8.9 & 79.5 & 59.7 & 11.2 & 28.9 & 83.3 & 63.8 \\
+ RFAConv & 4.0 & 10.5 & 80.6 & 60.7 & 14.4 & 34.9 & 83.5 & 64.2 \\
+ SGDM & 3.3 & 9.1 & $\mathbf{80.9}$ & $\mathbf{61.3}$ & 11.8 & 29.7 & $\mathbf{84.0}$ & $\mathbf{65.2}$ \\
\bottomrule
\end{tabular}
}
\end{table*}

\subsection{Implementation details}
\label{Implementation details}
To demonstrate that SGDM, as a plug-and-play module, can be easily inserted, we did not make significant changes to the structure of all base models. Simply put, SGDM is inserted into the above networks before the P3, P4, and P5 output layers before detection heads. The insertion method of SGDM and all other comparative models is consistent for all experiments. To ensure a fair comparison of experiments, we used identical parameters for each group of base models. In order to facilitate researchers to reproduce SGDM in the YOLO framework better, the core parameters are referenced in \cref{table1}. All of the experiments are based on a double NVIDIA RTX A4000 platform.

\subsection{Results Comparison on VOC}

\begin{table}[t]
\caption{We embed different dynamic convolutions in SGDM to verify the rationality of RDConv's design. For CondConv, DynamicConv, and ODConv, the important parameter "kernel number" is set to the default value of 4. We selected two backbones, YOLOv5n and YOLOv8n. The best results are bolded.}
\label{table3}
\centering
\begin{tabular}{lccccc}
\toprule
SGDM & Param(M) & FLOPs(G) & $\text{AP}_{50}(\%)$ & $\text{AP}(\%)$ & Time(ms)\\
\midrule
YOLOv5n &2.5 & 7.2 & 77.9 & 56.3 & 1.0 \\
+ CondConv & 3.5 & 9.0 & 79.3 & 60.0 & 1.2 \\
+ DynamicConv & 3.5 & 9.0 & 79.6 & 59.8 & 1.2 \\
+ DCD & 2.8 & 8.1 & 78.5 & 58.7 & 1.1 \\
+ ODConv & 3.5 & 9.0 & 79.2 & 59.5 & 1.3 \\
+ RDConv & 2.8 & 8.1 & $\textbf{79.9}$ & $\textbf{60.3}$ & \textbf{1.0}\\
\midrule
YOLOv8n & 3.1 & 8.7 & 79.4 & 59.3 & 1.0 \\
+ CondConv & 4.0 & 10.0 & 79.9 & 60.9 & 1.6 \\
+ DynamicConv & 4.0 & 10.0 & 80.4 & 61.1 & 1.6 \\
+ DCD & 3.3 & 9.1 & 79.8 & 60.6 & 1.4 \\
+ ODConv & 4.0 & 10.0 & 80.1 & 61.2 & 1.7 \\
+ RDConv & 3.3 & 9.1 & $\textbf{80.9}$ & $\textbf{61.3}$ & \textbf{1.0}\\
\bottomrule
\end{tabular}
\end{table}

As shown in \cref{table2}, for CBAM and CA, the same embedding method does not bring stable performance improvement. For example, after adding CBAM to YOLOv8n, the AP decreased by 0.4\%, and after adding CA to YOLOv6n, the AP decreased by 0.2\%. On the other hand, SGDM shows performance improvements in six different backbones. For Nano models, SGDM has an average of 1.2\% AP improvement compared to the second-best RFAConv; for Small models, SGDM has an average of 0.6\% AP improvement. 

In \cref{table3}, We used two models, YOLOv5n and YOLOv8n, with the same embedded SGDM method. We replaced RDConv with other dynamic weight convolutions inside the SGDM. we can see that RDConv in SGDM exhibits superior performance compared to mainstream dynamic weight convolutions while achieving nearly identical inference speed as the Baseline model when paired with RDConv. 

In addition to the embedded experiments on YOLO series models, we selected three currently popular backbones to compare embedding performance to demonstrate the excellent compatibility of SGDM with many object detection models. We do not change the embedding method in \cref{Implementation details}, embedding SGDM into the Neck structure. From \cref{table4}, embedding the SGDM module can effectively improve the performance of various object detection networks, proving that SGDM has a wide range of applications.

\begin{table}[h]
\caption{Performance comparison of multiple object detection backbones embedded with SGDM. The best results are bolded.}
\label{table4}
\centering
\begin{tabular}{lcccc}
\toprule
Backbones & Param(M) & FLOPs(G) & $\text{AP}_{50}(\%)$ & $\text{AP}(\%)$ \\
\midrule
EfficientViT-M0 & 4.0 & 9.5 & 78.0 & 58.3  \\
+ SGDM & 4.2 & 9.7 & \textbf{79.6} & \textbf{58.8}  \\
\midrule
FasterNet-T0 & 4.2 & 10.7 & 79.2 & 59.0  \\
+ SGDM & 4.4 & 11.1 & \textbf{79.9} & \textbf{59.5} \\
\midrule
ConvNeXtV2-A & 5.7 & 14.1 & 77.8 & 57.8  \\
+ SGDM & 2.7 & 6.2 & \textbf{78.9} & \textbf{57.9} \\
\bottomrule
\end{tabular}
\end{table}

\begin{figure*}[h]
\centering
\includegraphics[width=\linewidth]{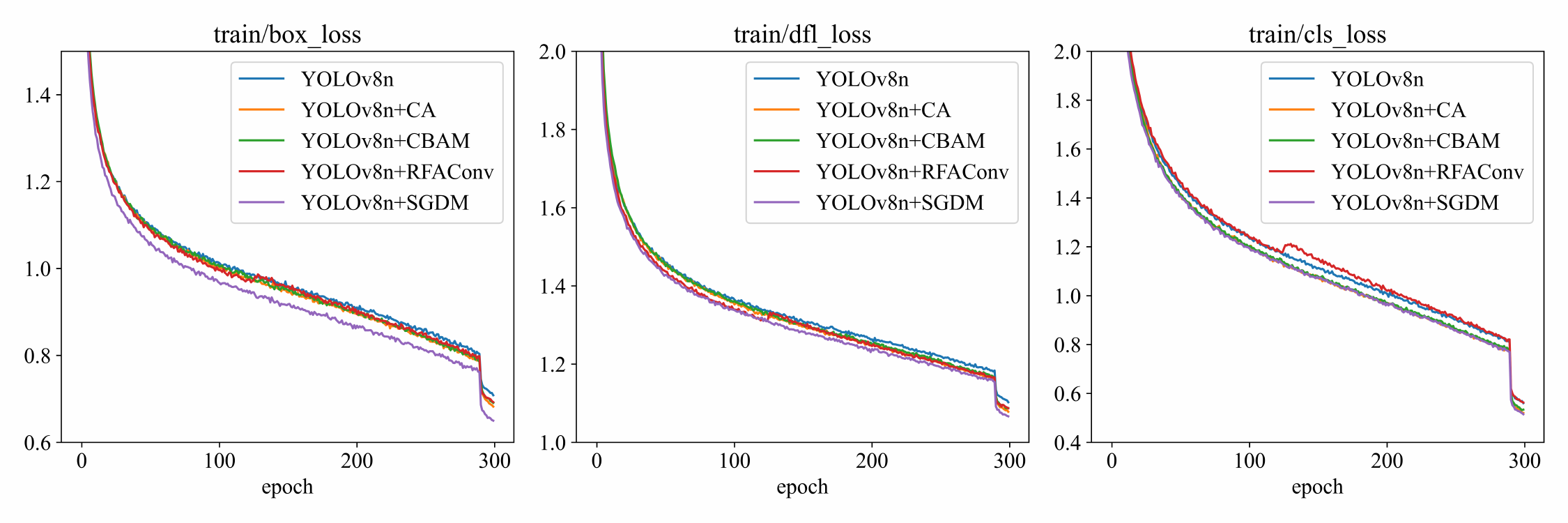}
\caption{The loss curve during the training process of YOLOv8n is embedded with four different attention mechanisms.}
\label{fig3}
\end{figure*}

Moreover, to better display the performance of different modules, we visualized the loss curve during the training process on YOLOv8n in \cref{fig3}. From the curves of three various losses, we can see that after embedding SGDM, the network can converge faster and better. At the 120th epoch, compared to RFAConv, which had the second-best performance, SGDM did not show any abnormal fluctuations, indicating that SGDM enhanced the robustness of the network.

\begin{table}[h]
\caption{$\text{AP}$, $\text{AP}_{75}$, $\text{AP}_{50}$, $\text{AP}_{S}$, $\text{AP}_{M}$, $\text{AP}_{L}$ results on the MS-COCO2017 validation sets. We also add a comparison of reasoning time. We adopt the YOLOv5n and YOLOv8n as baseline. Then, different plug-and-play attention modules are added, such as CA, CBAM, and RFAConv. The best results are bolded.}
\label{table5}
\centering
\begin{tabular}{lccccccccc}
\toprule
Backbones & Param(M) & FLOPs(G) & $\text{AP}$ & $\text{AP}_{75}$ & $\text{AP}_{50}$ & $\text{AP}_{S}$ & $\text{AP}_{M}$ & $\text{AP}_{L}$ & Time(ms)\\
\midrule
YOLOv5n & 1.9 & 4.6 & 27.5 & 28.9 & 45.6 & 13.5 & 31.5 & 35.9 & 4.4 \\
+ CA & 1.9 & 4.6 & 28.1 & 29.2 & 46.2 & 14.3 & 32 & 36.6 & 4.8 \\
+ CBAM & 1.9 & 4.6 & 27.6 & 28.6 & 45.5 & 13.6 & 31.2 & 36.6 & 5.4 \\
+ RFAConv & 2.7 & 6.2 & 29.0 & 30.6 & 47.3 & 14.8 & 33.4 & 37.4 & 5.3 \\
+ SGDM & 2.0 & 4.8 & $\textbf{30.1}$ & $\textbf{31.6}$ & $\textbf{47.9}$ & $\textbf{16.3}$ & $\textbf{34.2}$ & $\textbf{38.8}$ & $\textbf{5.2}$\\
\midrule
YOLOv8n & 3.1 & 8.7 & 36.4 & 39.7 & 51.9 & 18.4 & 40.1 & 52 & 4.2 \\
+ CA & 3.1 &8.8 & 36.7 & 39.9 & 52.1 & 17.8 & 40.3 & 51.6 & 4.5  \\
+ CBAM & 3.1 & 8.8 & 36.3 & 39.6 & 51.5 & 18.3 & 40.1 & 51.5 & 4.6 \\
+ RFAConv & 4.0 & 10.5 & 37.2 & 40.4 & 52.2 & 17.9 & 40.8 & $\textbf{53.5}$ & 4.5 \\
+ SGDM & 3.3 & 9.1 & $\textbf{38.1}$ & $\textbf{41.4}$ & $\textbf{54.0}$ & $\textbf{20.9}$ & $\textbf{42.2}$ & 51.3 & $\textbf{4.3}$\\
\bottomrule
\end{tabular}
\end{table}

\subsection{Results Comparison on MS-COCO} 
The results shown in \cref{table5} show similar performance improvement trends as on the VOC dataset. It can be seen that compared to other attention mechanisms, SGDM shows an AP improvement of 2.6\% and 1.7\% compared to the baseline models YOLOv5n and YOLOv8n, respectively. On YOLOv5n, after adding SGDM, the model achieved optimal results across all metrics with only a 5\% increase in parameters and a 4\% increase in computational complexity compared to the baseline model; additionally, it also demonstrated certain advantages in inference speed compared to adding other attention mechanisms. On YOLOv8n, except for $\text{AP}_{L}$ where its performance was not optimal, it outperformed various attention mechanisms across other metrics. Considering that RFAConv resulted in a 30\% increase in parameters and a 21\% increase in computational complexity, we can conclude that SGDM strikes a better balance between model accuracy and efficiency trade-off.

\begin{figure}[h]
\centering
\includegraphics[width=\linewidth]{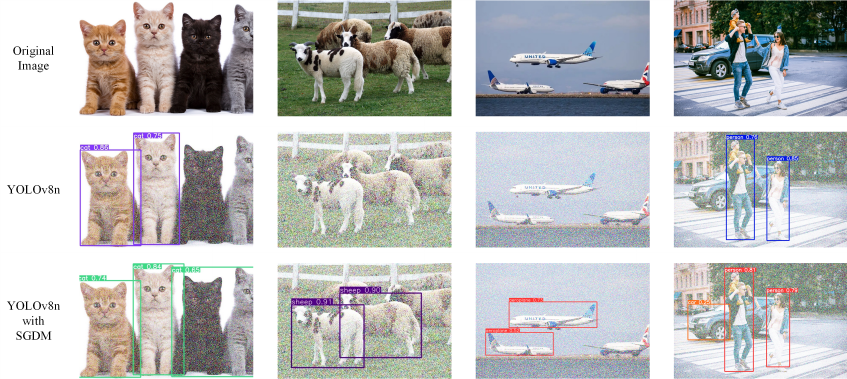}
\caption{Visualization of detection results for images with Gaussian noise using YOLOv8n model with and without SGDM.}
\label{fig4}
\end{figure}

Then, we verify that SGDM eliminates the problem of dynamic convolution being too sensitive to high-frequency noise through four sets of images in COCO. As shown in \cref{fig4}, we directly add global Gaussian noise to the images, which is unlikely to occur in reality but can better test the robustness of the network. After implementing SGDM, YOLOv8n significantly improves the detection rate and border accuracy. This proves that our RDConv effectively solves the problem of dynamic convolution being too sensitive to high-frequency noise.

\subsection{Ablation studies}
\label{Ablation studies}
We conducted multiple ablation experiments on the essential designs of RDConv and SGDM to demonstrate the rationality of parameter selection. All data are obtained by training YOLOv8n as the baseline model from scratch on the VOC dataset.

\paragraph{Feature Razor Ratio.}
Our first set of ablative experiments determines the feature razor ratio $r$ used in RDConv. From the results shown in \cref{table6}, we can find that under $r = 3/4$, $r = 1/2$, $r = 1/4$, and $r = 1/8$, SGDM can improve the performance of the baseline model. However, network performance has a process of rising and then falling, with the peak appearing at $r = 1/2$. When $r = 1/16$, there is a significant drop in network performance. We also added the measurement of parameter quantity and computational cost for a single SGDM module using a (2, 512, 40, 40) tensor. Finally, we chose $r=1/2$ as the parameter used in SGDM for a balance between accuracy improvement and efficiency.

\begin{table}[h]
\caption{Results comparison of different feature razor ratios $r$. The best results are bolded.}
\label{table6}
\centering
\begin{tabular}{lcccc}
\toprule
$r$ & Params(M) & FLOPs(G) & $\text{AP}_{50}(\%)$ & $\text{AP}(\%)$ \\
\midrule
Baseline & 3.1 & 8.7 & 79.4 & 59.3 \\
\midrule
$0.75$ & 0.78 & 2.22 & 80.3 & 61.2 \\
$0.5$ & 0.36 & 0.98 & \textbf{80.9} & \textbf{61.3} \\
$0.25$ & 0.10 & 0.25 & 79.3 & 59.8 \\
$0.125$ & 0.03 & 0.07 & 79.1 & 59.6 \\
$0.0625$ & 0.01 & 0.02 & 78.3 & 58.2 \\
\bottomrule
\end{tabular}
\end{table}

\paragraph{Kernel Size in Spatial Branch.}
As one of the cores of RDConv, the spatial branch is responsible for adding rich position information to generate dynamic kernel weight. Hence, choosing strip convolution kernel size in the spatial branch is particularly critical. In this group of ablation experiments, we selected $k_s=7$, $k_s=9$, $k_s=11$, $k_s=13$, and $k_s=15$ for comparison. Theoretically, the larger the convolution kernel, the better due to the increase of the receptive field. The results of \cref{table7} also confirm this point. Finally, to balance performance and computational complexity, we choose $k_s=15 $ as our default setting to balance parameter quantity and efficiency.

\paragraph{Static-Guided Feature Ratio.}
We conducted a series of ablation experiments on the design of SGDM as shown in \cref{table8}. As mentioned in the method section, we will divide the input features into four groups. Then, we input them into asymmetric static convolution groups and RDConv with residual links. Therefore, choosing the proportion of each input is crucial for constructing SGDM. Therefore, we fixed the number of channels entering the dynamic branch to equal that of two static branches, denoted as $r$. We chose four experimental groups with $r = 0.15$, $r = 0.2$, $r = 0.25$, and $r = 0.3$. 

We found that SGDM has optimal results using even distribution on the four branches. We speculated that combining our selection of feature razor ratio as 0.5 in RDConv, when $r$ is set to 0.25, the convolution kernels corresponding to channel numbers in two asymmetric static convolutions do not need to undergo Conv operations again but can be directly mapped to RDConv's weights, this operation contains the most original feature information. So $r=0 .25$ is also chosen as SGDM's final choice.

\begin{minipage}[t]{\linewidth}
\begin{minipage}[t]{0.48\linewidth}
\makeatletter\def\@captype{table}
\caption{Results comparison of different values of $k$ in spatial branch. The best results are bolded.}
\centering
\begin{tabular}{ccc}
\toprule
$k_s$ & $AP_{50}(\%)$ & $AP(\%)$\\
\midrule
$7$ & 80.6 & 60.8 \\
$9$ & 80.7 & 60.8 \\
$11$ & 80.9 & 60.9  \\
$13$ & \textbf{81.0} & 61.0 \\
$15$ & 80.9 & \textbf{61.3}\\
\bottomrule
\end{tabular}
\label{table7}
\end{minipage}
\begin{minipage}[t]{0.48\linewidth}
\makeatletter\def\@captype{table}
\caption{Results comparison of different group ratios for dynamic and static convolutions. The best results are bolded.}
\centering
\begin{tabular}{lcc}
\toprule
$r$ & $AP_{50}(\%)$ & $AP(\%)$\\
\midrule
$0.15$ & 80.5 & 61.2 \\
$0.2$ & 80.0 & 60.6 \\
$0.25$ & \textbf{80.9} & \textbf{61.3} \\
$0.3$ & 80.2 & 60.8 \\
\bottomrule
\end{tabular}
\label{table8}
\end{minipage}
\end{minipage}

\section{Conclusion}
In this paper, we propose RDConv and SGDM to combine the advantages of dynamic convolution with spatial attention mechanisms. We summarized the three problems that make dynamic convolution challenging to apply in spatial attention mechanisms. To remedy these problems, we use the razor operation to reduce the parameters during dynamic convolution execution, thereby significantly reducing the computational burden brought by introducing dynamic convolution, making it possible to design plug-and-play modules. Then, we design a spatial branch to overcome the neglect of spatial information in the dynamic convolution weight generation process, making dynamic convolutions more suitable for dense object tasks. Finally, our SGDM uses a pair of asymmetric static convolution parameters to guide the generation of dynamic convolution parameters, effectively overcoming the sensitivity to high-frequency noise introduced by dynamic convolutions. Since SGDM is a plug-and-play module, it can be easily deployed into existing networks without requiring major structural adjustments, further expanding the applicability of our module. Extensive experimental results show that SGDM has decent performance improvement in various datasets and backbones. Also, our embedding has a lower parameter count and inference speed loss than other mainstream attention mechanisms and dynamic convolutions, proving our design is effective.

\bibliographystyle{splncs04}
\bibliography{main}
\end{document}